\pdfoutput=1

\documentclass[11pt]{article}

\usepackage[preprint]{acl}

\usepackage{times}
\usepackage{latexsym}
\usepackage{tabularx}
\usepackage{longtable}
\usepackage{array}
\usepackage[justification=justified,singlelinecheck=false]{caption}

\usepackage[T1]{fontenc}

\usepackage[utf8]{inputenc}

\usepackage{microtype}

\usepackage{inconsolata}

\usepackage{CJKutf8}
\usepackage{amsmath}
\usepackage{hyperref}

\usepackage{graphicx}
\usepackage{booktabs}
\title{Quantifier Scope Interpretation in Language Learners and LLMs}

\author{
  Shaohua Fang\textsuperscript{1,2},
  Yue Li\textsuperscript{2},
  Yan Cong\textsuperscript{2,3} \\
  \textsuperscript{1}Department of English, Purdue University \\
  \textsuperscript{2}Department of Linguistics, Purdue University \\
  \textsuperscript{3}School of Languages and Cultures, Purdue University \\
  \texttt{\{fang413, li4207, cong4\}@purdue.edu}
}

\begin{document}
\maketitle
\begin{abstract}
Sentences with multiple quantifiers often lead to interpretive ambiguities, which can vary across languages. This study adopts a cross-linguistic approach to examine how large language models (LLMs) handle quantifier scope interpretation in English and Chinese, using probabilities to assess interpretive likelihood. Human similarity (HS) scores were used to quantify the extent to which LLMs emulate human performance across language groups. Results reveal that most LLMs prefer the surface scope interpretations, aligning with human tendencies, while only some differentiate between English and Chinese in the inverse scope preferences, reflecting human-similar patterns. HS scores highlight variability in LLMs’ approximation of human behavior, but their overall potential to align with humans is notable. Differences in model architecture, scale, and particularly models’ pre-training data language background, significantly influence how closely LLMs approximate human quantifier scope interpretations. 
\end{abstract}

\section{Introduction}

The interpretation of sentences containing two quantifiers—such as an existential quantifier and a universal quantifier—varies cross-linguistically. For example, the doubly quantified (DQ) sentence “A child climbed every tree” exhibits scope ambiguity. Under the surface scope (SS) reading, a single child climbed multiple trees, whereas the inverse scope (IS) reading suggests that different trees were climbed by different children. Similarly, when the quantifiers’ order is reversed, as in “Every child climbed a tree,” the resulting sentence in English also remains ambiguous. In this case, the surface scope reading implies that each child climbed a different tree, while the inverse scope reading suggests that all the children climbed the same tree. In Mandarin Chinese (henceforth Chinese), however, the prevailing theoretical view is that only the surface scope reading is permitted \citep{aoun1989scope, huang1998logical, lee1986studies}.

A substantial body of empirical evidence indicates that surface scope readings are generally preferred over inverse scope readings in English \citep{kurtzman1993resolution, lidz2018scope}. This preference is evident in several areas: surface scope readings are acquired earlier by children \citep{lidz2002children, gennari2006acquisition}, more easily acquired by second language (L2) learners \citep{wu2019l1, chu2014acquisition}, and processed more efficiently by native speakers \citep{anderson2004structure, brasoveanu2015strategies, dotlavcil2015manner}. One notable explanation for the increased difficulty associated with inverse scope is that its processing incurs a higher cognitive cost due to the complex syntactic operations required for derivation via covert movement at Logical Form, due to the Processing Scope Economy (PSE) principle proposed by \citet{anderson2004structure}. This principle predicts that IS should incur greater processing cost, as its derivation involves greater structural complexity due to covert movement, which creates a mismatch between surface syntax and semantics—unlike SS. Compared to English, there is considerably less empirical evidence on scope interpretation in Chinese. Findings from a number of experimental studies present a more nuanced picture regarding scope rigidity in Chinese, with some research suggesting that inverse scope readings may be available to Chinese speakers \citep{fang2023quantifier, fang2025experimental, scontras2017cross, zhou2009bscope}.

Thus far, psycholinguistic approaches to quantifier scope interpretation have provided significant insights into both the grammatical representation and cognitive processing of scope phenomena. More recently, the advent of large language models (LLMs), has sparked growing interest in their ability to handle a variety of linguistic phenomena, such as interference effects \citep{cong2023language, li2025beyond} and discourse connectives \citep{britton2024influence}. However, limited attention has been given to how LLMs interpret scope relations, a process that involves complex cognitive interactions between syntax and semantics. Pertaining to our work, \citet{kamath2024scope} focused exclusively on the “every...a” configuration in English DQ sentences. To expand this line of inquiry, the current study adopts a cross-linguistic approach to investigate how LLMs process scope interpretations of doubly quantified sentences with different syntactic configurations in both English and Chinese, as in Experiment 1. The study also compares LLM performance with human data on quantifier scope interpretation to assess similarities and differences in their interpretative patterns, as in Experiment 2.

\section{Experiment 1}
\subsection{Stimuli and dataset}

We adopted a truth-value judgment paradigm commonly used in scope studies \citep{ionin2010scope, zhou2009ascope}, designing 60 experimental target sentences per language. These included 30 existential quantifier (UE)  and 30 reverse order (EU) sentences, adapted from \citet{fang2023quantifier}. Each sentence was paired with two story contexts: one favoring a surface scope (SS) interpretation and one favoring an inverse scope (IS) interpretation. For example, the SS context for “Every child climbed a tree” described three children each climbing a different tree during a playground game. 

The Chinese version was a direct translation of the English materials, including both the target sentences and their associated interpretations. For instance, “Every child climbed a tree” was translated as \begin{CJK}{UTF8}{gbsn}“每一个孩子都爬了一棵树”\end{CJK}. See Table~\ref{tab:gloss_ch} for a glossed analysis. All translations were reviewed by the first author, a native Mandarin speaker and trained linguist, to ensure naturalness and linguistic appropriateness. Each language set consisted of 120 trials (60 UE, 60 EU). Example sentences and interpretations are provided in Appendix A.

\begin{table}[h!]
\centering
{\fontsize{10}{10.5}\selectfont
\setlength{\tabcolsep}{3pt} 
\begin{tabular}{lllllll}
\hline
\begin{CJK}{UTF8}{gbsn}每一个\end{CJK} & 
\begin{CJK}{UTF8}{gbsn}孩子\end{CJK} & 
\begin{CJK}{UTF8}{gbsn}都\end{CJK} & 
\begin{CJK}{UTF8}{gbsn}爬了\end{CJK} & 
\begin{CJK}{UTF8}{gbsn}一\end{CJK} & 
\begin{CJK}{UTF8}{gbsn}棵\end{CJK} & 
\begin{CJK}{UTF8}{gbsn}树\end{CJK} \\
měi yí-gè & háizi & dōu & pá-le & yì & kē & shù \\
every one-CL & child & all & climb-ASP & one & CL & tree \\
\hline
\end{tabular}
}
\vspace{2pt}

\caption{Gloss of the Chinese sentence \begin{CJK}{UTF8}{gbsn}“每一个孩子都爬了一棵树”\end{CJK} (‘Every child climbed a tree’).}
\label{tab:gloss_ch}
\end{table}

\subsection{Models and computational approach}

We tested seven language models: BERT-family: BERT-base (110M) \citep{RN40}, BERT-large (340M) \citep{RN40}; GPT-family: DistilGPT2 (82M) \citep{RN52}, GPT-2 trained on English (GPT-2En, 124M) \citep{radford2019language}, and GPT-2 trained on Chinese (GPT-2Ch, 95M) \citep{zhao2019uer, zhao2023tencentpretrain}; LLaMA-family: LLaMA models trained on English (LlamaEn, 7B) \citep{touvron2023llama} and Chinese (LlamaCh, 7B) \citep{cui2023efficient}. 



Interpretation preference was assessed via the conditional probability of a sentence being accepted in a given context, following the rationale of truth-value judgment tasks. For example, acceptance of “Every child climbed a tree” in an IS-favoring context implies access to the IS interpretation. The conditional probability is computed by the conditional score function using the Minicons library \citep{misra2022minicons}.

For autoregressive models (e.g., GPT, LLaMA), sentence probabilities were computed by multiplying token-level probabilities via the chain rule. For masked models (e.g., BERT), we used pseudo-log-likelihood scoring, masking each token sequentially and aggregating prediction probabilities, which were then transformed to surprisals.



Lower surprisal indicates a more expected (preferred) interpretation. Each sentence was assigned a binary label: 1 if the SS interpretation had lower surprisal (i.e., was preferred), and 0 if the IS interpretation was preferred.

To examine interpretation preferences, we fit logistic mixed-effects models with interpretation (SS = 1, IS = 0) as the dependent variable, and language as a fixed effect. In parallel, surprisal scores were analyzed using linear mixed-effects models to evaluate the influence of language, interpretation, and LLM. All predictors were sum-coded. Random intercepts were included for items. $p$-values were estimated using the \texttt{lmerTest} package \citep{kuznetsova2017lmertest}, and post-hoc comparisons were conducted with \texttt{emmeans} \citep{lenth2020estimated} using Tukey adjustment.

\subsection{Hypotheses}

Our general hypothesis is that if LLMs' behavior is consistent and interpretable with respect to a certain scope processing pattern (c.f., PSE), as seen in humans, then we would expect the alignment of LLMs and humans in our scope interpretation experiments. Taking humans' behavior as an interpretation baseline, we measure the degree to which LLMs' behavior can be consistently interpreted by human sentence processing theories. We examined multiple LLMs to tease apart how sophisticated linguistic phenomena such as DQ were represented and processed in LLMs' pre-training, ultimately improving LLMs' transparency. Concretely, for both UE and EU constructions, we predict that IS interpretation would impose greater processing costs than SS for LLMs, similar to human participants, despite variations in LLM performance due to differences in scale and architecture.

\subsection{Results}

\begin{figure}[htbp]
  \centering
  \includegraphics[width=0.5\textwidth]{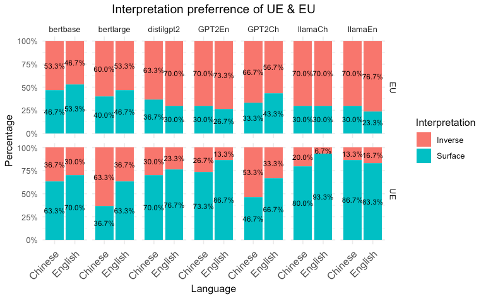}
  \caption{LLMs’ preferred interpretations (surface vs. inverse) by structure (UE vs. EU) and language (English vs. Chinese). Note: “Surface” = SS interpretation; “Inverse” = IS interpretation.}
  \label{fig:mean-percentages}
\end{figure}

Figure~\ref{fig:mean-percentages} illustrates the distribution of interpretation preferences across different LLMs for UE and EU structures in both English and Chinese. For UE structures, SS readings were generally preferred over IS readings across LLMs, except for BERT-large and GPT2Ch in the Chinese dataset.  we conducted a mixed-effects regression analysis with Language as a predictor for each LLM. The analysis revealed a significant main effect of Language for BERT-large only ($b = 1.1,\ p = .0499$), indicating that IS interpretations were more likely in Chinese than in English. For EU structures, all LLMs predominantly preferred IS readings, except for BERT-base in the English dataset. This preference for IS readings remained consistent across English and Chinese, as indicated by the absence of significant Language effect across LLMs (all $ps > 0.5$).
We conducted statistical comparisons between SS and IS for each LLM within each language using numerical surprisal values. 

\begin{figure}[htbp]
  \centering
  \includegraphics[width=0.5\textwidth]{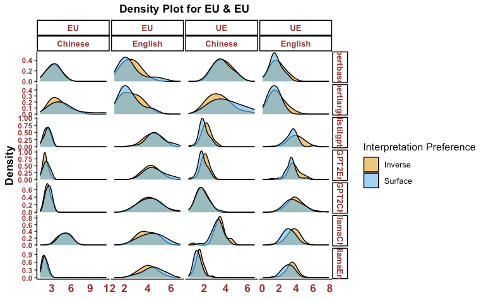}
  \caption{LLM surprisal distributions for surface vs. inverse interpretations in UE and EU structures (English vs. Chinese).}
  \label{fig:mean-surprisal}
\end{figure}

The descriptive results were visualized in Figure~\ref{fig:mean-surprisal} for both structures. Tables~\ref{tab:UE-summary} and~\ref{tab:EU-summary} present the surprisal values for both SS and IS across structures, where lower surprisal scores indicate a higher likelihood of a given reading. As shown in Table~\ref{tab:UE-summary} from a series of linear mixed-effects models for the UE structure, most LLMs demonstrated sensitivity to the distinction between SS and IS, with a general preference for SS over IS. An exception was BERT-large for Chinese in the UE condition, where IS emerged as the more likely reading. In the case of EU as shown in Table~\ref{tab:EU-summary}, fewer LLMs exhibited the ability to differentiate between SS and IS compared to UE. Notably, among the models that significantly distinguished between SS and IS, IS was the more likely reading for both English and Chinese, mirroring the patterns observed in the binary categorical data.

Given that English and Chinese arguably differ in the extent to which IS is permitted, we filtered the surprisal values for IS across both languages for each LLM. Language (Chinese and English), LLM, and their interaction were included as independent variables to examine how the likelihood of permitting IS varies by language and model. In the case of UE structures, it is unsurprising that both Language ($b = 0.6, \ p < .001$) and LLM ($b = -0.3,\ p < .001$) returned significant main effects. Of particular interest was the significant interaction obtained between Language and LLM ($b = -2.5,\ p < .001$). Post hoc pairwise comparisons revealed some interesting contrasts: with the exception of LlamaCH, all LLMs showed significant differences between Chinese and English in their likelihood of allowing IS, albeit in different directions (all $ps < .001$). Specifically, for BERT-base and BERT-large, IS was more likely in English than in Chinese. In contrast, for the other LLMs (DistilGPT2, GPT-2En, GPT-2Ch, LlamaEn), IS was more likely in Chinese than in English.

For the EU structure, a linear mixed-effects model with Language and LLM as fixed effects indicated significant main effects for both Language ($b = 0.8,\ p < .001$) and LLM ($b = 1.1,\ p < .001$). More importantly, a significant interaction between Language and LLM was observed ($b = -1.5,\ p < .001$). Post hoc pairwise comparisons demonstrated that IS was more likely in English than in Chinese for BERT-base, BERT-large, and LlamaCH (all $ps < .001$). In contrast, for the remaining four LLMs, IS was more likely in Chinese than in English (all $ps < .001$).

\begin{table}[htbp]
\centering
\scriptsize
\resizebox{\linewidth}{!}{%
\begin{tabular}{@{}lccc|ccc@{}}
\toprule
\multicolumn{1}{c}{} & \multicolumn{3}{c|}{\textbf{English}} & \multicolumn{3}{c}{\textbf{Chinese}} \\
\textbf{Model} & \textbf{SS} & \textbf{IS} & \textbf{Sig.} & \textbf{SS} & \textbf{IS} & \textbf{Sig.} \\
\midrule
BERT-base   & 1.6 & 1.8 & **  & 3.7 & 3.7 & .16 \\
BERT-large  & 1.4 & 1.7 & **  & 4.0 & 3.4 & *   \\
DistilGPT2  & 3.5 & 4.0 & *** & 2.0 & 2.2 & **  \\
GPT-2En     & 3.4 & 3.8 & *** & 1.9 & 2.1 & **  \\
GPT-2Ch     & 3.9 & 4.0 & .23 & 1.9 & 1.8 & .29 \\
LlamaEn     & 3.1 & 3.4 & *** & 1.4 & 1.6 & *** \\
LlamaCh     & 3.2 & 3.6 & *** & 3.2 & 3.4 & **  \\
\bottomrule
\end{tabular}%
}
\scriptsize\textit{Note:} Significance levels of $p$ values: *$p<.05$, **$p<.01$, ***$p<.001$
\vspace{2pt}
\caption{Mean surprisal and statistical comparison of SS vs. IS by LLM and Language in UE structure}
\raggedright
\label{tab:UE-summary}
\end{table}

\begin{table}[htbp]
\centering
\scriptsize
\resizebox{\linewidth}{!}{%
\begin{tabular}{@{}lccc|ccc@{}}
\toprule
\multicolumn{1}{c}{} & \multicolumn{3}{c|}{\textbf{English}} & \multicolumn{3}{c}{\textbf{Chinese}} \\
\textbf{Model} & \textbf{SS} & \textbf{IS} & \textbf{Sig.} & \textbf{SS} & \textbf{IS} & \textbf{Sig.} \\
\midrule
BERT-base   & 2.6 & 2.6 & .98 & 3.3 & 3.3 & .95 \\
BERT-large  & 2.5 & 2.6 & .56 & 4.9 & 4.0 & *   \\
DistilGPT2  & 4.7 & 4.4 & *   & 2.3 & 2.3 & .35 \\
GPT-2En     & 4.6 & 4.3 & **  & 2.2 & 2.0 & .05 \\
GPT-2Ch     & 4.2 & 4.2 & .81 & 2.5 & 2.4 & .19 \\
LlamaEn     & 4.4 & 4.0 & **  & 2.0 & 1.8 & **  \\
LlamaCh     & 4.3 & 4.0 & *   & 5.1 & 4.9 & .16 \\
\bottomrule
\end{tabular}%
}
\scriptsize\textit{Note:} Significance levels of $p$ values: *$p<.05$, **$p<.01$, ***$p<.001$
\vspace{2pt}
\caption{Mean surprisal and statistical comparison of SS vs. IS by LLM and Language in EU structure}
\raggedright
\label{tab:EU-summary}
\end{table}

\section{Experiment 2}

In addition to evaluating the performance of various LLMs on linguistic tasks—specifically, quantifier scope interpretation, which entails the complex interplay between syntax and semantics—this study also aims to investigate the extent to which LLMs demonstrate human-like linguistic generalizations, a crucial question that has garnered growing attention in recent research \citep{cai2024large}; \cite{hu2024language}; \cite{dentella2023systematic}. In Experiment 2, we address this issue by comparing LLMs’ scope interpretations with human judgments. 

\subsection{Overview of human data}
The human data was taken from a previously published dataset \citep{fang2023quantifier}, which includes participants from four groups: first language (L1) English native speakers, L1 Mandarin Chinese native speakers, L2 English speakers with L1 Chinese, and L2 Chinese speakers with L1 English\footnote{Native English speakers were recruited via \href{https://www.prolific.com}{Prolific}. Other groups were drawn from populations in the US, UK, and Mainland China, depending on language background.}. The data was collected using Truth Value Judgment Tasks (TVJTs), a format that our Experiment 1 was designed to replicate.

\textbf{Experiment} Participants completed a Truth Value Judgment Task (TVJT), in which they rated the acceptability of sentences embedded in story contexts that supported either a surface scope (SS) or inverse scope (IS) interpretation. Ratings were provided on a 7-point Likert scale. L1 and L2 English speakers completed the English version of the task, while L1 and L2 Chinese speakers completed the Chinese version translated from the original English materials. Each DQ configuration (UE, EU) included 12 critical sentences, with two potential readings per sentence, resulting in 24 items per structure. 

\textbf{Findings} The main findings of \citet{fang2023quantifier} are as follows: (1) IS interpretations were more difficult than SS in English but still more acceptable than their Chinese counterparts; and (2) L2 groups, on the whole, were able to acquire the target interpretation.

\subsection{Quantifying human–LLM alignment via human similarity score}

To quantify the alignment between LLM outputs and human judgments, we adopted the Human Similarity (HS) Score proposed by \citet{duan2024hlb}. This metric evaluates how closely LLM response distributions match human responses on individual test items using Jensen-Shannon (JS) divergence, a symmetric, bounded measure derived from Kullback-Leibler (KL) divergence. As such, a lower JS divergence results in a higher HS score, indicating that the LLM’s responses are more human-like. 





\begin{figure}[htbp]
  \centering
  \includegraphics[width=0.48\textwidth]{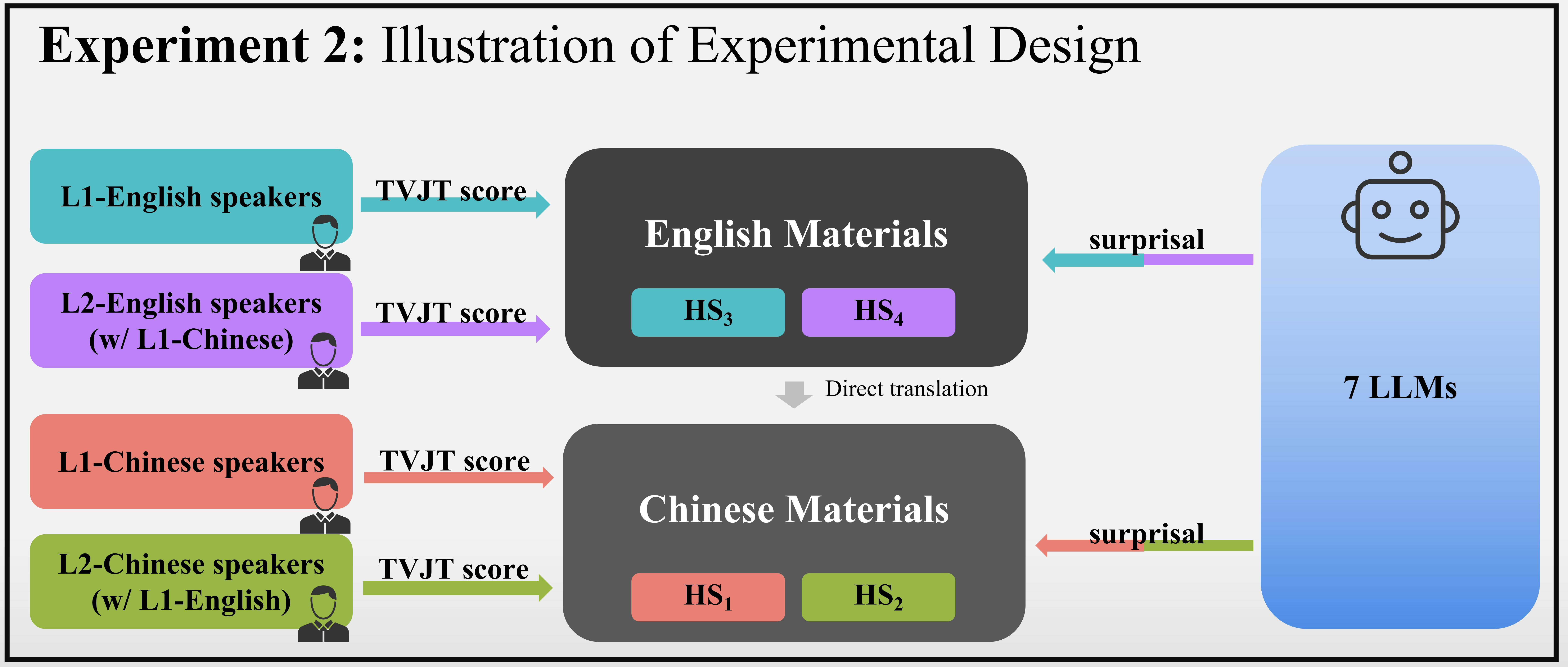}
  \caption{Overview of Experimental 2 design, illustrating how surprisal-based results from LLMs are compared against human judgments from four participant groups using matched materials. Each Human Similarity (HS) score reflects the alignment between LLM response distributions (from surprisal values) and human ratings (from Truth Value Judgment Tasks) for a specific group-material pair. HS$_1$–HS$_4$ correspond to: HS$_1$: L1-Chinese speakers on Chinese materials vs. LLMs; HS$_2$: L2-Chinese speakers (L1-English) on Chinese materials vs. LLMs; HS$_3$: L1-English speakers on English materials vs. LLMs; HS$_4$: L2-English speakers (L1-Chinese) on English materials vs. LLMs.}
  \label{fig:exp2-overview}
\end{figure}

A note that we consider the human similarity score as a descriptive metric to better understand similarities and differences of LLMs and humans in our quantifier scopes tasks. As a single composite score, we maintain that human similarity \citet{duan2024hlb} provides a gradient proxy to more precisely index human-LLMs alignment. See Figure~\ref{fig:exp2-overview} for experiment 2 illustration.

\subsection{Results}

\begin{figure}[htbp]
  \centering
  \includegraphics[width=0.5\textwidth]{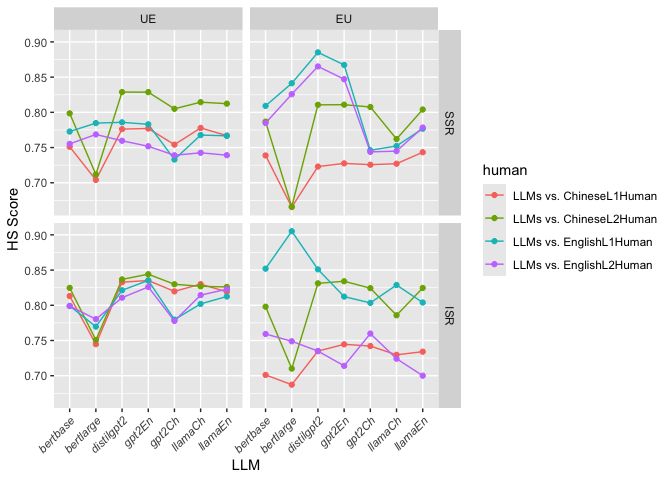}
    \caption{HS scores of LLMs for UE and EU sentences under each interpretation (SS/IS), based on comparisons between LLM surprisals and human TVJT responses. Red and green lines represent HS scores on Chinese materials: red for native Chinese speakers (L1-Chinese), green for L2 Chinese learners with L1 English. Blue and purple lines represent HS scores on English materials: blue for L1 English speakers, purple for L2 English learners with L1 Chinese. For example, the red line in the UE-SS panel (top left) shows the similarity between LLMs and L1 Chinese speakers on SS interpretations of UE sentences.}
  \label{fig:humanlikeness}
\end{figure}

Figure~\ref{fig:humanlikeness} presents the aggregated HS across individual items, resulting from the comparison between LLM predictions and human judgments for UE and EU scope interpretations, respectively. 

As shown in the figure, HS scores revealed significant variations in how LLMs resembled human language use. Several notable patterns emerged: First, for UE sentences (left panels), LLMs demonstrated the highest performance when compared to L2 Chinese learners for human similarity, while for EU (right panels), the models overall were least similar to human performance when compared to Chinese L1 speakers and most similar when compared to English L1 speakers. HS scores for comparisons between Chinese L1 speakers and LLMs tend to be low  across both UE and EU constructions. Second, among the models, BERT-large showed the weakest performance, particularly when compared to Chinese L1 and L2 speakers, while the GPT family consistently performed the best, and the Llama family performed moderately. Third, as for the potential influence of training data’s language, GPT2CH and LlamaCH performed better with UE constructions for Chinese L1 and Chinese L2 speakers than for English L1 and L2 speakers across both SS and IS readings, whereas for EU constructions, English L1 and Chinese L2 speakers outperformed the other two language groups.

We conducted a series of ANOVA tests to compare the performance of different model families for each quantifier scope construction across language groups. The interpretations were aggregated for each construction. For UE, post hoc comparisons revealed that GPT ($p = .022$) and Llama ($p = .076$) outperformed BERT when comparing LLM performance to L2 Chinese speakers. For EU, GPT ($p = .027$) and Llama ($p = .039$) also outperformed BERT when compared to Chinese L1 speakers. Additionally, for EU and LLM performance compared to L2 Chinese speakers, GPT outperformed BERT ($p = .016$). In a nutshell, BERT seemed to exhibit the worst performance in light of handling quantifier scope in a way comparable to human participants particularly when the human data came from Chinese native speakers or learners of Chinese. 

A two-factor ANOVA revealed a significant interaction between the language group and LLM on HS scores for the EU construction only but not for the UE construction. Post hoc pairwise comparisons showed that, with BERT, HS scores were significantly higher for English L1 than for Chinese L1 ($p < .0001$), higher for English L2 than for Chinese L1 ($p = .023$), higher for English L1 than for Chinese L2 ($p = .001$), and marginally higher for English L1 than for English L2 ($p = .053$). With GPT, HS scores for Chinese L2 ($p = .002$) and English L1 ($p = .0006$) were significantly higher than for Chinese L1. Given the significant results for GPT, the influence of its subtypes was analyzed, but no interaction was found between GPT subtypes and language groups.

\section{Discussion}

\subsection{Semantic representation \& interpretation in LLMs}

Despite English’s interpretive ambiguity and its differences from Chinese, SS was preferred over IS due to lower processing costs \citep{anderson2004structure}, PSE principle). LLMs favored SS, especially in UE constructions (Table ~\ref{tab:UE-summary}. However, cross-linguistic variation in IS acceptance received limited support, with only BERT-base and BERT-large showing lower IS likelihood in Chinese than in English. 

The results from LLMs for UE constructions, specifically the strong preference for SS over IS readings, align with human data reported by \citet{fang2023quantifier} and 
\citet{scontras2017cross}. However, these studies also revealed clear cross-linguistic patterns, showing that Chinese does not permit IS as readily as English, as evidenced by lower human ratings for IS compared to SS. In this study, BERT models were the most consistent and interpretable: only data from the BERT family models corroborated the cross-linguistic variation, mirroring the directional patterns observed in human participants. 

LLM results for EU sentences were unexpected. While human data favored SS over IS in both English and Chinese, LLMs showed the opposite pattern. Like UE constructions, BERT models preferred IS more in English than Chinese, partially aligning with human data. LlamaCH matched human patterns, whereas other models diverged, exhibiting a reversed trend.

In psycholinguistic experiments with human participants, EU constructions in English exhibit a sharper preference for SS over IS compared to UE constructions. This contrasts with the results from the LLM experiments in our study. Established explanations account for this human behavior, two of which are particularly noteworthy. The first explanation is that UE constructions inherently entail SS, meaning the IS reading describes a specific situation already covered by the SS reading. As a result, IS is easier to access in UE constructions compared to EU constructions, where such entailment does not occur \citep{reinhart1976syntactic}; \cite{ruys2002wide}. As shown in Tables 1 and 2, this phenomenon is presumably linked to the higher likelihood of IS in UE than in EU, at least in English. The second explanation stems from the Single Reference (SR) Principle by \citet{kurtzman1993resolution}, a processing principle suggesting that listeners construct an online parse of sentences incrementally. When encountering a singular indefinite at the beginning of a sentence, listeners tend to commit to a single referent associated with it. Consequently, EU constructions with a singular subject are strongly biased toward SS, making IS even harder to access than in UE constructions. Given these explanations, the performance of LLMs on both construction types suggests they could capture subtle aspects of semantics, such as entailment, and processing principles like the PSE Principle, mirroring patterns observed in human participants. 

In comparison, \citet{kamath2024scope}, one of the very few studies investigating LLMs’ treatment of scopally ambiguous sentences, also provided evidence supporting preferences for inverse scope readings over surface scope readings by LLMs, consistent with human preferences, specifically for UE constructions in English. This aligns, to a large extent, with the data patterns observed in our study for UE constructions. Notably, \citet{kamath2024scope} employed a different testing format, presenting a test sentence alongside two explicit statements, each corresponding to a particular interpretation. This indicates that direct queries may yield results comparable to those derived from surprisal-based data collection, as in our approach. That said, \citet{kamath2024scope} exclusively tested UE constructions. Consequently, any findings regarding these constructions, whether from humans or LLMs, should be interpreted with caution due to the aforementioned potential entailment issues, which may undermine the validity of assessing the availability of inverse scope and comparing it to surface scope. In this sense, our study, which includes both EU and UE constructions and examines Chinese data in addition to the English data studied by \citet{kamath2024scope}, represents a novel contribution to the field. 

Experiment 1 enables a descriptive comparison of LLM and human performance in psycholinguistic tasks, while Experiment 2 quantifies their alignment using HS scores. Numerical comparisons are crucial, as cognitive modeling with LLMs aims to illuminate natural language processing. Extending \citet{duan2024hlb}, we explore quantifier scope, a syntactic-semantic challenge, to assess LLMs’ capacity for linguistic complexity.

\subsection{LLMs approximate ``human-like'' language use but with greater variability}

Our results from the HS scores suggest that LLMs generally performed well in approximating human language use, with most language groups achieving scores exceeding 0.7—higher than those reported by \citet{duan2024hlb} across tasks probing various linguistic levels. However, similar to \citet{duan2024hlb}, considerable variation in HS scores was observed across different language groups, with GPT and Llama models outperforming BERT models. This result can be explained by the architectures of these LLMs. GPT and Llama, as autoregressive decoder models, inherently align better with next-word prediction tasks measured with surprisals \citep{tunstall2022natural}. 

In contrast, BERT, as an encoder-based model, is less well-suited for modeling quantifier scope interpretation based on surprisals. Another plausible explanation is model scale. While larger models generally yield better performance, this assumption does not fully account for the differences observed between GPT, Llama, and BERT. Despite BERT’s average parameter sizes being larger than GPT’s as in Table 1, its performance was comparatively worse. This finding supports the inverse scaling hypothesis \citep{mckenzie2023inverse}, which posits that LLM performance may decrease with increasing model scale, as observed in other linguistic phenomena as well, such as semantic attraction \citep{cong2023language}.

That said, BERT models appeared to outperform other models in revealing contrastive patterns for IS in English versus Chinese. As evidenced in Experiment 1, only BERT models demonstrated patterns resembling human data, showing that IS was more likely in English than in Chinese for both UE and EU constructions. The role of the training data’s language emerged in this study, albeit with mixed evidence. For EU constructions, LlamaCh applied to both English and Chinese data exhibited a significantly lower surprisal score for Chinese than for English with IS, suggesting that LlamaCh may have captured from its pre-training data the unlikelihood of IS arising in Chinese. However, GPT2Ch did not replicate this pattern, indicating that the influence of training language may not consistently impact model behavior. 

The effect of training language was further highlighted in the interaction between human language group and LLM. BERT, with a pretraining corpus primarily composed of English texts, performed better in handling quantifier scope for English, even when the output involved English L2 learners. Conversely, GPT’s modeling results showed weaker approximations for Chinese L1 compared to Chinese L2 and English L1. This suggests that GPT is better equipped to handle English data (from English L1) and Chinese data with potential English features (from Chinese L2). Overall, these findings highlight the role of pretraining data’s language in shaping LLM performance. These results point towards the limited multilingual capacities of LLMs, particularly when their pretraining data is dominated by monolingual sources \citep{zhang2023don}.

\subsection{To some extent, LLM is more like L2}

Most LLM studies focus on monolingual L1 data, making our inclusion of L2 data a novel contribution. For each construction, we compared LLM performance for English L1 and English L2 speakers, and Chinese L1 and Chinese L2. The results revealed that for EU, LLMs aligned more closely with English L1 than English L2 speakers, particularly for the IS reading. In the case of UE, regardless of SS or IS, LLMs received higher HS scores when compared to Chinese L2 learners than to Chinese L1 speakers. Similarly, for EU, LLM performance more closely resembled that of Chinese L2 than Chinese L1 speakers.

Overall, our findings indicate that LLMs exhibit linguistic preferences similar to L1-English, L2-Chinese human participants. As previously argued, LLM-generated text more closely resembles English native speaker productions or Chinese texts written by native English speakers learning Chinese, likely reflecting linguistic transfer from English. Our findings add to and resonate with the previous research by \cite{schut2025multilingual}, suggesting that these models perform core reasoning and representational processing in a space fundamentally shaped by English. 

An important question is whether this alignment is influenced by the training language of LLMs. To examine this, we analyzed models trained primarily in Chinese (e.g., GPT2Ch and LlamaCh). Statistical analyses revealed no significant interaction between language group (L1 vs. L2) and training language, suggesting that even LLMs trained in Chinese align more closely with L1-English, L2-Chinese participants. We hypothesize that this effect may stem from the presence of translated English content in the training data of these Chinese-trained LLMs, which could contribute to their linguistic similarity with English-dominant patterns.

As an exploratory analysis, we tested the more recent and powerful Deepseek-R1 \citep{guo2025deepseek}, in a zero-shot setting (API: https://api-docs.deepseek.com) using the same dataset to evaluate interpretation preferences across contexts. Our prompt instructed the model: “\textit{You are an AI model trained to compute conditional probabilities. Given the {context}, what is the probability that the following {target sentence} will occur next? Respond with ONLY a number from 0.0 to 1.0 and nothing else.}” Deepseek-R1 showed an overall strong preference for SS readings, selecting it in 96.7\% of Chinese cases and over 73\% of English cases. IS was more acceptable in English than in Chinese. Overall, Deepseek-R1’s performance closely mirrored that of other tested LLMs, particularly in its strong preference for SS over IS. However, it appears more promising in capturing cross-linguistic variation in the extent to which IS is permitted in each language. We leave more systematic studies of deepseek-r1 for future research.

\subsection{Quantifier scope as a diagnostic of semantic transparency in LLMs}

Our findings suggest that quantifier scope offers a uniquely revealing test of whether LLMs represent meaning beyond surface statistics. Unlike many syntactic tasks, scope resolution involves latent semantic interpretation and interaction with context, making it a high-precision probe of how robustly LLMs represent linguistic meaning. Critically, scope preferences vary across languages. This makes cross-linguistic scope behavior an ideal benchmark for semantic transparency. Our findings suggest partial transparency. BERT models, in particular, approximate cross-linguistic contrasts in IS availability. However, more recent models such as Llama fail to replicate these contrasts, despite scoring higher on global human similarity metrics. This disconnect indicates that surface fluency does not imply robust semantic understanding.

Interestingly, the LLMs' behavior also parallels that of L2 learners. Like L2 speakers, LLMs tend to default to structurally simpler, surface-scope interpretations and show limited flexibility when faced with cross-linguistic semantic contrasts. We speculate that this pattern likely reflects pre-training biases analogous to L1 transfer in human learners. We leave systematic examination for future research.

In sum, scope ambiguity serves both as a syntactic-semantic challenge and as a window into how LLMs represent (and process) meaning. It distinguishes models that merely simulate ``human-like'' outputs from those that exhibit interpretive depth aligned with linguistic knowledge. As such, scope is a critical benchmark for assessing true linguistic competence in LLMs.

Although our methodology and results do not exhaust the most recent or state-of-the-art language models currently available, we maintain that our approach is generalizable for newer models within the same model family as those studied here. The diagnostic datasets and evaluation metrics we employ are designed to be broadly applicable. We hope that our experiments can continue to support meaningful analysis and comparison as new models in the same line are developed.

\section{Conclusion}

To conclude, our research demonstrates the capacity of LLMs to understand quantifier scope cross-linguistically. Most LLMs showed a preference for SS interpretations, similar to human participants, while only a subset of LLMs distinguished between English and Chinese in the differential likelihood of IS, reflecting human-like patterns. Although HS scores varied in the degree to which LLMs approximated human participants, they demonstrated remarkable human-like potential overall. Moreover, the pretraining data’s language background played a significant role in shaping the extent to which LLMs resemble humans from language groups in quantifier scope interpretation. Future research could scale up psycholinguistic experiments involving both human participants and LLMs to achieve more robust results. Additionally, multiple measures, including techniques from prompt engineering alongside surprisal, could be employed to cross-validate and triangulate the experimental findings.

\section{Limitations}

We acknowledge several limitations in this study that future research could address. First, the items used for deriving LLM data were not identical to those used in the human experiment: while the LLM evaluation items were expansions of those used with human participants, enriched in both quantity and contextual detail, this discrepancy may weaken the validity of direct comparisons. Future studies should consider using identical items for both human participants and LLMs to enable more reliable comparisons. That said, the Human Similarity metric employed in Experiment 2 does not strictly require identical items across human and LLM experiments.

Second, our study relies exclusively on surprisal values to evaluate LLM interpretation. Future research should incorporate additional metrics, such as prompt-based probing, to triangulate findings and enhance the generalizability of results. For instance, in our exploration with Deepseek, we prompted LLMs using detailed story contexts and asked them to judge the fit between sentences and contexts. This approach helps to provide more insights into how LLMs understand quantifier scope. Probing is especially valuable given that surprisal alone cannot determine whether a particular interpretation (e.g., inverse scope) is grammatically permitted in Chinese; it merely reflects relative likelihoods. In contrast, prompting with Likert scales (e.g., 7-point ratings) allows researchers to evaluate acceptability more directly. For example, across all Chinese items with inverse scope, if the average LLM rating were 2, compared to 6 for surface scope, it would suggest that inverse scope may not be licensed in Chinese grammar—at least in contrast to surface scope as the baseline.

\bibliography{ref,custom}

\appendix

\section{Appendix}
This appendix provides the full set of sample stimuli used in the experiment, organized by language (English, Chinese) and quantifier scope structure (UE, EU). Each row contains the test sentence and two interpretive contexts: one supporting the surface scope reading (SS) and one supporting the inverse scope reading (IS). These examples illustrate the stimuli design used in the truth-value judgment task. The full table begins on the next page.

\label{sec:appendix}

\renewcommand{\arraystretch}{1.3}

\begin{table*}[h]
\centering
\begin{tabularx}{\textwidth}{|p{1.5cm}|p{1.5cm}|p{2.5cm}|X|}
\hline
\textbf{Language} & \textbf{Structure} & \textbf{Example sentence} & \textbf{Interpretation} \\
\hline

English & UE & Every child climbed a tree. & \textbf{SS}:  One day, the three children played on the playground and decided to play a game to see who could climb to the top of the tree as soon as possible. There are three tall trees on the playground, and the height of each tree is almost the same. For the fairness of the game, they decided to choose a different tree to climb. After the start of the game, the children worked hard to climb up. They were very focused and wanted to be the first child to climb to the top of the tree. After some efforts, every child climbed to the top of the tree smoothly to celebrate his achievements excitedly. Every child climbed a tree and successfully completed the game. 

\textbf{IS}: One day, the three children decided to play a game to see who could climb to the top of the tree as soon as possible. There is only one big tree on the playground, so they decided to take turns climbing and record the time everyone spent. After the start of the game, the children tried one after another, and everyone tried their best to climb to the top of the tree in the shortest time. In the end, all children successfully completed the challenge and used different times. They excitedly discussed the results of the competition and celebrated their results.
Every child climbed a tree. \\
\hline

English & EU & A child climbed every tree. & \textbf{SS}:In this school, a boy particularly likes to climb trees. There are three tall trees on the playground. One day, he decided to challenge himself to see if he could successfully climb all the trees. So he started from the first tree, then climbed the second tree, and finally climbed the third tree. After some efforts, he successfully climbed into three trees and was very proud.

\textbf{IS}: In this school, three children particularly like to climb trees.There are three tall trees on the playground. One day, they decided to play a game to see who climbed the fastest.After the start of the game, the children quickly climbed up. In the end, each child climbed to the top of the tree. The results of the game were very fierce, and everyone was proud of their performance. \\
\hline

Chinese & UE & 
\begin{CJK}{UTF8}{gbsn} 每一个孩子都爬了一棵树。\end{CJK} & \textbf{SS}:\begin{CJK}{UTF8}{gbsn}有一天，三个孩子在操场上玩耍，决定进行一场比赛，看看谁能最快地爬到树顶。操场上有三棵高大的树，每棵树的高度都差不多相同。为了比赛公平，他们决定每人选择一棵不同的树来爬。比赛开始后，孩子们奋力向上攀爬，他们都非常专注，想成为第一个爬到树顶的孩子。经过一番努力，最终每个孩子都顺利地爬到了树顶，兴奋地庆祝自己的成就。每一个孩子都爬了一棵树，圆满完成了比赛。\end{CJK}

\textbf{IS}: \begin{CJK}{UTF8}{gbsn}有一天，三个孩子决定进行一场比赛，看看谁能最快爬到树顶。操场上只有一棵大树，因此他们决定轮流攀爬，记录下每个人所花的时间。比赛开始后，孩子们一个接一个地尝试，每个人都尽全力想要在最短的时间内爬到树顶。最后，所有孩子都顺利地完成了挑战，并分别用时不同。他们在操场上兴奋地讨论比赛的结果，庆祝各自的成绩。\end{CJK} \\
\hline

\end{tabularx}
\caption{Examples and interpretations by language and structure}
\end{table*}

\clearpage
\noindent
\begin{minipage}{\textwidth}
\begin{tabularx}{\textwidth}{|p{1.5cm}|p{1.5cm}|p{2.5cm}|X|}
\multicolumn{4}{l}{\textbf{Appendix (continued)}} \\
\hline
\textbf{Language} & \textbf{Structure} & \textbf{Example sentence} & \textbf{Interpretation} \\
\hline

Chinese & EU & 
\begin{CJK}{UTF8}{gbsn} 有一个孩子爬了每一棵树。\end{CJK} & \textbf{SS}:\begin{CJK}{UTF8}{gbsn}在这所学校里，有一个男孩特别喜欢爬树。操场上有三棵高大的树，有一天，他决定挑战自己，看看能否成功爬上所有的树。于是，他从第一棵树开始，接着爬第二棵，最后爬上了第三棵。经过一番努力，他成功地爬上了三棵树，感到非常骄傲。\end{CJK}

\textbf{IS}: \begin{CJK}{UTF8}{gbsn}在这所学校里，有三个孩子特别喜欢爬树。操场上有三棵高大的树，有一天，他们决定进行一场比赛，看看谁爬得最快。比赛开始后，孩子们迅速向上攀爬，最终每个孩子都顺利爬到了树顶，比赛结果非常激烈，大家都为自己的表现感到自豪。\end{CJK} \\
\hline

\end{tabularx}
\end{minipage}

\end{document}